\documentclass[runningheads]{llncs}
\usepackage[utf8]{inputenc}
\usepackage{graphicx}
\usepackage{amssymb,amsmath,mathtools}
\usepackage{enumitem}
\usepackage[pdftex, pdfborder={0 0 0}]{hyperref}
\usepackage[textsize=small]{todonotes}
\usepackage{pgfplots}
\usepackage{tikz}
\usepackage{csquotes}
\usepackage{marvosym}
\usepackage{subcaption}

\newcommand{\cosim}{\operatorname{sim}} %
\newcommand{\ve}[1]{\mathbf{#1}} %
\newcommand{\tsum}{\sum\nolimits}

\newcommand{\norm}[1]{\left\lVert#1\right\rVert}

\newcommand{\sprod}[2]{\left<#1,#2\mathstrut\right>}

\newcommand{\refsec}[1]{Section~\ref{#1}}
\newcommand{\reffig}[1]{Fig.~\ref{#1}}
\newcommand{\reftab}[1]{Table~\ref{#1}}
\newcommand{\refeqn}[1]{Eq.~\refeq{#1}}

\begin{document}

\title{A Triangle Inequality for Cosine Similarity%
\thanks{Part of the work on this paper has been supported by Deutsche Forschungsgemeinschaft (DFG),
project number 124020371,
within the Collaborative Research Center SFB 876
``Providing Information by Resource-Constrained Analysis'', project A2 
}}
\author{Erich~Schubert}
\titlerunning{A Triangle Inequality for Cosines}
\authorrunning{E. Schubert}
\institute{TU Dortmund University, Dortmund, Germany
\\ 
\email{erich.schubert@tu-dortmund.de}
}
\maketitle

\begin{abstract}
Similarity search is a fundamental problem for many data analysis techniques.
Many efficient search techniques rely on the triangle inequality of metrics,
which allows pruning parts of the search space based on transitive bounds on distances.
Recently, Cosine similarity has become a popular alternative choice to the standard
Euclidean metric, in particular in the context of textual data and neural network embeddings.
Unfortunately, Cosine similarity is not metric and does not
satisfy the standard triangle inequality. Instead, many search techniques
for Cosine rely on approximation techniques such as locality sensitive hashing.
In this paper, we derive a triangle inequality for Cosine similarity that is
suitable for efficient similarity search with many standard search structures
(such as the VP-tree, Cover-tree, and M-tree); show that this bound is tight
and discuss fast approximations for it.
We hope that this spurs new research on accelerating exact similarity search
for cosine similarity, and possible other similarity measures
beyond the existing work for distance metrics.
\end{abstract}

\section{Introduction}

Similarity search is a fundamental problem in data science and is used as a building
block in many tasks and applications, such as nearest-neighbor classification,
clustering, anomaly detection, and of course information retrieval.
A wide class of search algorithms requires a metric distance function, i.e., a
dissimilarity measure $d(x,y)$ that satisfies the triangle inequality
$d(x,y) \leq d(x,z) + d(z,y)$ for any $z$. Intuitively, this is the requirement
that the direct path from $x$ to $y$ is the shortest, and any detour over another
point $z$ is at least as long. Many dissimilarity measures such as the popular
Euclidean distance and Manhattan distance satisfy this property, but not all do:
for example the \emph{squared} Euclidean distance
(minimized, e.g., by the popular $k$-means clustering algorithm)
does not, even on univariate data: $d^2_{\text{Euclid}}(0,2)=2^2=4$
but $d^2_{\text{Euclid}}(0,1)+d^2_{\text{Euclid}}(1,2)=1^2+1^2=2$.

The triangle inequality $d(x,y) \leq d(x,z) + d(z,y)$ is a central technique in
accelerating similarity search because it allows us to compute a bound on a
distance $d(x,y)$ without having to compute it exactly if we know $d(x,z)$ and
$d(z,y)$ for some object $z$.
With trivial rearrangement and relabeling of $y$ and $z$, we can also obtain
a lower bound: $d(x,y) \geq d(x,z) - d(z,y)$.
Given a maximum search radius $\varepsilon$, if $d(x,z) - d(z,y)>\varepsilon$,
we can then infer that $y$ cannot be part of the search result.
This technique is often combined with a search tree, where each subtree $Z$ is associated
with a routing object $z$, and stores the maximum distance $d_{\max}(z):=\max_{y\in Z} d(z,y)$ for all
$y$ in the subtree. If $d(x,z)-d_{\max}(z)>\varepsilon$, none of the objects in
the subtree can be part of the search result, and we can hence skip many candidates
at once.
This technique can also be extended to $k$-nearest neighbor and priority search,
where we can use the minimum possible distance $d(x,z)-d_{\max}(z)$ to prioritize
or prune candidates.

Metric similarity search indexes using this approach include
the ball-tree~\cite{tr/icsi/Omohundro89},
the metric tree~\cite{DBLP:journals/ipl/Uhlmann91} aka.{} the vantage-point tree~\cite{DBLP:conf/soda/Yianilos93},
the LAESA index~\cite{DBLP:journals/prl/MicoOV94,DBLP:conf/sisap/RuizSCFT13},
the Geometric Near-neighbor Access Tree (GNAT)~\cite{DBLP:conf/vldb/Brin95} aka.{} multi-vantage-point-tree~\cite{DBLP:journals/tods/BozkayaO99},
the M-tree~\cite{DBLP:conf/vldb/CiacciaPZ97},
the SA-tree~\cite{DBLP:journals/vldb/Navarro02} and Distal SAT~\cite{DBLP:conf/sisap/ChavezLRR14},
the iDistance index~\cite{DBLP:journals/tods/JagadishOTYZ05},
the cover tree~\cite{DBLP:conf/icml/BeygelzimerKL06},
the M-index~\cite{DBLP:journals/is/NovakBZ11},
and many more. (Neither the k-d-tree, quad-tree, nor the R-tree belong to this family, these indexes
are coordinate-based, and require lower-bounds based on hyperplanes and bounding boxes, respectively.)
While they differ in the way they organize the data (e.g., with nested balls in the ball tree, M-tree, and cover tree,
by splitting into ball-and-remainder in the VP-tree, or by storing the distances to reference
points in LAESA and iDistance), all of these examples rely on the triangle inequality for pruning candidates
as central search technique, and should not be used with a distance that does not satisfy this condition,
as the search results may otherwise be incomplete (this may, however, be acceptable for certain applications).

In this paper, we introduce a triangle inequality for Cosine similarity that allows lifting most of these
techniques from metric distances to Cosine similarity, and we hope that future research will allow extending
this to other popular similarity functions.

\section{Cosine Distance and Euclidean Distance}

Cosine similarity (which we will simply denote as ``$\cosim$'' in the following)
is commonly defined as the Cosine of the angle $\theta$ between two vectors $\ve{x}$ and $\ve{y}$:
$$
\cosim(\ve{x},\ve{y}):=
\operatorname{sim}_{\text{Cosine}}(\ve{x},\ve{y}) :=
\frac{\sprod{\ve{x}}{\ve{y}}}{\norm{\ve{x}}_2\cdot \norm{\ve{y}}_2}
=\frac{\sum_i x_iy_i}{\sqrt{\sum_i x_i^2}\cdot \sqrt{\sum_i y_i^2}}
=\cos \theta
$$
Cosine similarity has some interesting properties that make it a popular choice
in certain applications, in particular in text analysis.
First of all, it is easy to see that $\cosim(\ve{x},\ve{y})=\cosim(\alpha\ve{x},\ve{y})=\cosim(\ve{x},\alpha\ve{y})$
for any $\alpha>0$, i.e., the similarity is invariant to scaling vectors with a positive scalar.
In text analysis, this often is a desirable property as repeating the contents of a document multiple times
does not change the information of the document substantially.
Formally, Cosine similarity can be seen as being the dot product of $L_2$ normalized vectors.
Secondly, the computation of Cosine similarity is fairly efficient for sparse vectors:
rather than storing the vectors as a long array of values,
most of which are zero, they can be encoded for example as pairs $(i,v)$ of an index $i$
and a value $v$, where only the non-zero pairs are stored and kept in sorted order.
The dot product of two such vectors can then be efficiently computed by a \emph{merge} operation,
where only those indexes $i$ need to be considered that are in both vectors because in
$\sprod{\ve{x}}{\ve{y}}=\sum_i x_i y_i$ only those terms matter where both $x_i$ and $y_i$
are not zero.

In popular literature, you will often find the claim that Cosine similarity is
more suited for high-dimensional data. As we will see below, it cannot be superior to
Euclidean distance because of the close relationship of the two, hence this must be
considered a myth. Research on intrinsic dimensionality has shown that
Cosine similarity is also affected by the distance concentration effect~\cite{DBLP:conf/recsys/NanopoulosRI09}
as well as the hubness phenomenon \cite{DBLP:conf/icml/RadovanovicNI09},
two key aspects of the ``curse of dimensionality''~\cite{DBLP:journals/sadm/ZimekSK12}.
The main difference is that we are usually using the Cosine similarity on sparse
data, which has a much lower intrinsic dimensionality than the vector space dimensionality suggests.

Consider the Euclidean distance of two \emph{normalized} vectors $\ve{x}$ and $\ve{y}$.
By expanding the binomials, we obtain:
\begin{align}
d_{\text{Euclidean}}(\ve{x},\ve{y}) :=&
\sqrt{\tsum_i (x_i - y_i)^2}
=
\sqrt{\tsum_i (x_i^2 + y_i^2 - 2x_iy_i)}
\notag
\\
=&
\sqrt{\norm{\ve{x}}^2+\norm{\ve{y}}^2 -2 \sprod{\ve{x}}{\ve{y}}}
=
\sqrt{\sprod{\ve{x}}{\ve{x}}+\sprod{\ve{y}}{\ve{y}} -2 \sprod{\ve{x}}{\ve{y}}}
\label{eqn:euclid-scalar}
\\&
\text{if $\norm{\ve{x}}{=}\norm{\ve{y}}{=}1$: }
=\sqrt{2 - 2 \cdot \cosim(\ve{x},\ve{y})}
\label{eqn:euclid}
\end{align}
where the last step relies on the vectors being normalized to unit length.
Hence we have an extremely close relationship between Cosine similarity and
(squared) Euclidean distance of the normalized vectors:
\begin{align}
\cosim(\ve{x},\ve{y}) &= 1 - \tfrac12 d_{\text{Euclidean}}^2\big(\tfrac{\ve{x}}{\norm{\ve{x}}},\tfrac{\ve{y}}{\norm{\ve{y}}}\big) \enskip.
\label{eq:cosim-euclid}
\end{align}

While we can also compute Euclidean distance more efficiently for sparse vectors
using the scalar product form of \refeqn{eqn:euclid-scalar},
this computation is prone to a numerical problem called ``catastrophic cancellation''
for small distances (when $\sprod{\ve{x}}{\ve{x}}\approx\sprod{\ve{x}}{\ve{y}}\approx\sprod{\ve{y}}{\ve{y}}$
that can be problematic in clustering (see, e.g., \cite{DBLP:conf/ssdbm/SchubertG18,DBLP:conf/sisap/LangS20}).
Hence, working with Cosines directly is preferable when possible, and additional motivation
for this work was to work directly with a triangle inequality on the similarities,
to avoid this numerical problem (as we will see below, we cannot completely avoid this,
unless we can afford to compute many trigonometric functions).

In common literature, the term ``Cosine distance'' usually refers to a dissimilarity function defined as
\begin{align}
d_{\text{Cosine}}(\ve{x},\ve{y}) &:= 1 - \cosim(\ve{x},\ve{y}) \enskip,
\label{eq:cosdist1}
\\
\shortintertext{which unfortunately is \emph{not} a metric, i.e., it does not satisfy the
triangle inequality. There are two less common alternatives, namely:}
d_{\text{SqrtCosine}}(\ve{x},\ve{y}) &:= \sqrt{2 - 2\cosim(\ve{x},\ve{y})}
\enskip
\Big(= d_{\text{Euclidean}}\big(\tfrac{\ve{x}}{\norm{\ve{x}}},\tfrac{\ve{y}}{\norm{\ve{y}}}\big)
\enskip\Big)
\label{eq:cosdist2}
\\
d_{\text{arccos}}(\ve{x},\ve{y}) &:= \arccos(\cosim(\ve{x},\ve{y})) \enskip.
\label{eq:cosdist3}
\end{align}
which are less common (but, e.g., available in ELKI~\cite{DBLP:journals/corr/abs-1902-03616})
and which are metric. \refeqn{eq:cosdist2} directly follows from \refeqn{eq:cosim-euclid},
while the second one is the angle between the vectors itself (the arc length, not the cosine of the angle),
for which we easily obtain the triangle inequality by looking at the arc through $x$, $y$, $z$.
We will use these metrics below to obtain a triangle inequality for Cosines.

\section{Constructing a Triangle Inequality for Cosine Similarity}

Because the triangle inequality is the central rule to avoiding distance computations
in many metric search indexes (as well as in many other algorithms), we would like
to obtain a triangle inequality for Cosine similarity.
Given the close relationship to squared Euclidean distance outlined in the previous section,
one obvious approach would be to just use Euclidean distance instead of Cosine.
If we know that our data is normalized (which is a best practice when using Cosine similarities),
we can make the computation slightly more efficient using \refeqn{eq:cosdist2},
but we wanted to avoid this because
(i)~computing the square root takes 10--50 CPU cycles (depending on the exact CPU, precision, and input value)
and (ii)~the subtraction in this equation is prone to catastrophic cancellation when the two vectors are similar,
i.e., we may have precision issues when finding the nearest neighbors.
Hence, we would like to develop techniques that primarily rely on similarity instead of distance,
yet allow a similar pruning to the (very successful) metric search acceleration techniques.

Using \refeqn{eq:cosdist2} and the triangle inequality of Euclidean distance, we obtain
\begin{align}
\sqrt{1 - \cosim(\ve{x},\ve{y})}
&\leq
\sqrt{1 - \cosim(\ve{x},\ve{z})} + \sqrt{1 - \cosim(\ve{z},\ve{y})}
\notag
\\
\cosim(\ve{x},\ve{y})
&\geq
1-\big(\sqrt{1 - \cosim(\ve{x},\ve{z})} + \sqrt{1 - \cosim(\ve{z},\ve{y})}\big)^2
\notag
\\
\cosim(\ve{x},\ve{y})
&\geq
\cosim(\ve{x},\ve{z}) + \cosim(\ve{z},\ve{y}) - 1
\notag\\
&\phantom{{}\geq{}}- 2\sqrt{\big(1 - \cosim(\ve{x},\ve{z}))(1 - \cosim(\ve{z},\ve{y})\big)}
\label{eq:bound-eucl-1}
\shortintertext{
which, unfortunately, does not appear to allow much further simplification.
In order to remove the square root, we can approximate it using the smaller of the two similarities
$\cosim_\bot(\ve{x},\ve{y},\ve{z}):=\min\{\cosim(\ve{x},\ve{z}), \cosim(\ve{z},\ve{y})\}$:
}
\cosim(\ve{x},\ve{y})
&\geq
\cosim(\ve{x},\ve{z}) + \cosim(\ve{z},\ve{y}) - 1 - 2(1-\cosim_\bot(\ve{x},\ve{y},\ve{z}))
\notag
\\
\cosim(\ve{x},\ve{y})
&\geq
\cosim(\ve{x},\ve{z}) + \cosim(\ve{z},\ve{y}) + 2 \cosim_\bot(\ve{x},\ve{y},\ve{z}) - 3
\label{eq:bound-eucl-2}
\end{align}
This is highly efficient to compute, a strict bound to \refeqn{eq:bound-eucl-1}, but unfortunately also a rather loose bound if one
of the similarities is high, but the other is not.

Besides the relationship to squared Euclidean distance, there is another way to obtain a
metric from Cosine similarity, namely by using the arc length as in \refeqn{eq:cosdist3}
(i.e., using the angle $\theta$ itself, rather than the Cosine of the angle):
\begin{align}
d_{\arccos}(\ve{x},\ve{y}) &:= \arccos(\cosim(\ve{x},\ve{y}))
\notag
\shortintertext{
This also yields a metric on the sphere that satisfies the triangle inequality:
}
\arccos(\cosim(\ve{x},\ve{y})) &\leq \arccos(\cosim(\ve{x},\ve{z})) + \arccos(\cosim(\ve{z},\ve{y}))
\notag
\shortintertext{and, hence,}
\cosim(\ve{x},\ve{y}) &\geq
\cos(\arccos(\cosim(\ve{x},\ve{z})) + \arccos(\cosim(\ve{z},\ve{y})))
\label{eq:bound-acos-1}
\shortintertext{
Computationally, the trigonometric functions involved here are even more expensive (60--100 CPU cycles each),
hence using this variant directly is not for free. However, this can be
further transformed (c.f., angle addition theorems) to the following equivalent triangle inequality for Cosine similarity:
}
\cosim(\ve{x},\ve{y}) &\geq \cosim(\ve{x},\ve{z})\cdot \cosim(\ve{z},\ve{y})
\notag
\\&\phantom{={}}
- \sqrt{(1-\cosim(\ve{x},\ve{z})^2)\cdot (1-\cosim(\ve{z},\ve{y})^2)}
\enskip.
\label{eq:bound-acos-2}
\end{align}
This triangle inequality is \emph{tighter} than the one based on Euclidean distance,
and hence we can expect better pruning power than using an index for Euclidean distance
or $d_{\text{SqrtCosine}}$ (\refeqn{eq:cosdist2}) in a metric index; while the computational cost
has been reduced to the low ``overhead'' of Euclidean distances.
\refeqn{eq:bound-acos-1} suggests that it is the tightest possible bound we can obtain because
it is directly using the angles, rather than the chord length as used by Euclidean distance.
This bound yields a very interesting insight: while the triangle inequality for Euclidean distances
-- and in the arc lengths -- was additive, the main term of this equation in the Cosine domain is multiplicative.

We also investigated approximations to further reduce the computation overhead.
By approximating the last term using the smaller similarity only, we get
\begin{align}
\cosim(\ve{x},\ve{y}) &\geq
\cosim(\ve{x},\ve{z})\cdot \cosim(\ve{z},\ve{y}) + \min\{\cosim(\ve{x},\ve{z})^2,\cosim(\ve{z},\ve{y})^2\} - 1
\label{eq:bound-acos-3}
\end{align}
which is a cheap bound, tighter than \refeqn{eq:bound-eucl-2}, but still too loose. 

We can also expand and approximate the last term using both the smaller and the larger value
$\cosim_\top(\ve{x},\ve{y},\ve{z}):=\max\{\cosim(\ve{x},\ve{z}), \cosim(\ve{z},\ve{y})\}$:
\begin{align*}
&\phantom{{}={}}\sqrt{(1-\cosim(\ve{x},\ve{z})^2)\cdot (1-\cosim(\ve{z},\ve{y})^2)}
\notag\\
&=
\sqrt{(1-\cosim(\ve{x},\ve{z}))\cdot (1+\cosim(\ve{x},\ve{z}))\cdot (1-\cosim(\ve{z},\ve{y})) \cdot (1+\cosim(\ve{z},\ve{y}))}
\notag\\
&\leq \sqrt{(1+\cosim_\top(\ve{x},\ve{y},\ve{z}))^2\cdot (1-\cosim_\bot(\ve{x},\ve{y},\ve{z}))^2}
\notag\\
&= (1+\cosim_\top(\ve{x},\ve{y},\ve{z}))\cdot (1-\cosim_\bot(\ve{x},\ve{y},\ve{z}))
\notag\\
&= 1+\cosim_\top(\ve{x},\ve{y},\ve{z})-\cosim_\bot(\ve{x},\ve{y},\ve{z})-\cosim(\ve{x},\ve{z})\cdot \cosim(\ve{z},\ve{y})
\notag
\end{align*}
and hence obtain the inequality
\begin{align}
\cosim(\ve{x},\ve{y}) &\geq 2\cdot \cosim(\ve{x},\ve{z})\cdot \cosim(\ve{z},\ve{y})
- 1 - |\cosim(\ve{x},\ve{z}) - \cosim(\ve{z},\ve{y})| 
\label{eq:bound-acos-4}
\end{align}
but this approximation is strictly inferior to \refeqn{eq:bound-acos-3}.

\subsection{Opposite Direction}

The opposite direction of the triangle inequality is often as important
as the first direction. For distances and the angles, it is simply obtained
by moving one term to the other side and renaming. It can then be simplified
as before
\begin{align}
\arccos(&\cosim(\ve{x},\ve{y})) \geq \arccos(\cosim(\ve{x},\ve{z})) - \arccos(\cosim(\ve{z},\ve{y}))
\notag
\\
\cosim(\ve{x},\ve{y}) &\leq \cos(\arccos(\cosim(\ve{x},\ve{z})) - \arccos(\cosim(\ve{z},\ve{y})))
\notag
\\
\cosim(\ve{x},\ve{y}) &\leq \cosim(\ve{x},\ve{z})\cdot\cosim(\ve{z},\ve{y})
+ \sqrt{(1-\cosim(\ve{x},\ve{z})^2)\cdot(1-\cosim(\ve{z},\ve{y})^2)}
\label{eq:bound-acos-rev}
\intertext{
It is interesting to see that Equations~\refeq{eq:bound-acos-2} and \refeq{eq:bound-acos-rev}
together imply that
}
|\cosim(\ve{x},\ve{y}) &- \cosim(\ve{x},\ve{z})\cdot\cosim(\ve{z},\ve{y})|
\leq
\sqrt{(1-\cosim(\ve{x},\ve{z})^2)\cdot(1-\cosim(\ve{z},\ve{y})^2)}
\notag
\end{align}
i.e., a symmetric error bound for $\cosim(\ve{x},\ve{y}) \approx \cosim(\ve{x},\ve{z})\cdot\cosim(\ve{z},\ve{y})$.

\section{Experiments}

\begin{table}[b!]
\caption{Triangle inequalities/bounds compared}
\label{tab:bounds}
\setlength{\tabcolsep}{5pt}
\begin{tabular}{lcl}
Name & Eq. & Equation \\
Euclidean & \eqref{eq:bound-eucl-1} &
$\cosim(\ve{x},\ve{z}) + \cosim(\ve{z},\ve{y}) - 1
- 2\sqrt{\big(1 - \cosim(\ve{x},\ve{z}))(1 - \cosim(\ve{z},\ve{y})\big)}$
\\
Eucl-LB & \eqref{eq:bound-eucl-2} &
$\cosim(\ve{x},\ve{z}) + \cosim(\ve{z},\ve{y}) + 2 \cdot \min\{\cosim(\ve{x},\ve{z}),\cosim(\ve{y},\ve{z})\} - 3$
\\
\bf Arccos & \eqref{eq:bound-acos-1} &
$\cos(\arccos(\cosim(\ve{x},\ve{z})) + \arccos(\cosim(\ve{z},\ve{y})))$
\\
\bf Mult & \eqref{eq:bound-acos-2} &
$\cosim(\ve{x},\ve{z})\cdot \cosim(\ve{z},\ve{y}) - \sqrt{(1-\cosim(\ve{x},\ve{z})^2)\cdot (1-\cosim(\ve{z},\ve{y})^2)}$
\\
Mult-LB1 & \eqref{eq:bound-acos-3} &
$\cosim(\ve{x},\ve{z})\cdot \cosim(\ve{z},\ve{y}) + \min\{\cosim(\ve{x},\ve{z})^2,\cosim(\ve{y},\ve{z})^2\} - 1$
\\
Mult-LB2 & \eqref{eq:bound-acos-4} &
$2\cdot \cosim(\ve{x},\ve{z})\cdot \cosim(\ve{z},\ve{y})
- |\cosim(\ve{x},\ve{z}) - \cosim(\ve{z},\ve{y})| - 1$
\end{tabular}
\end{table}

\reftab{tab:bounds} summarizes the six bounds that we compare concerning their suitability for metric indexing.
Note that we will not investigate the actual performance in a similarity index here,
but plan to do this in future work. Instead, we want to focus on the bounds themselves concerning three properties:

\begin{enumerate}
\item how tight the bounds are, i.e., how much pruning power we lose
\item whether we can observe numerical instabilities
\item the differences in the computational effort necessary 
\end{enumerate} 

\begin{figure}[tb]\centering
\includegraphics[width=.8\linewidth]{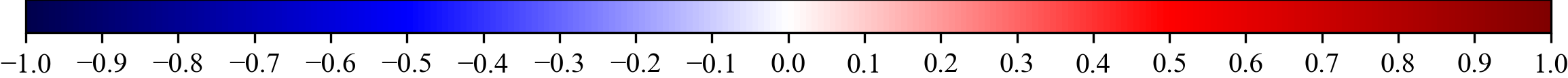}
\\
\begin{subfigure}{.3\linewidth}\centering
\includegraphics[height=35mm]{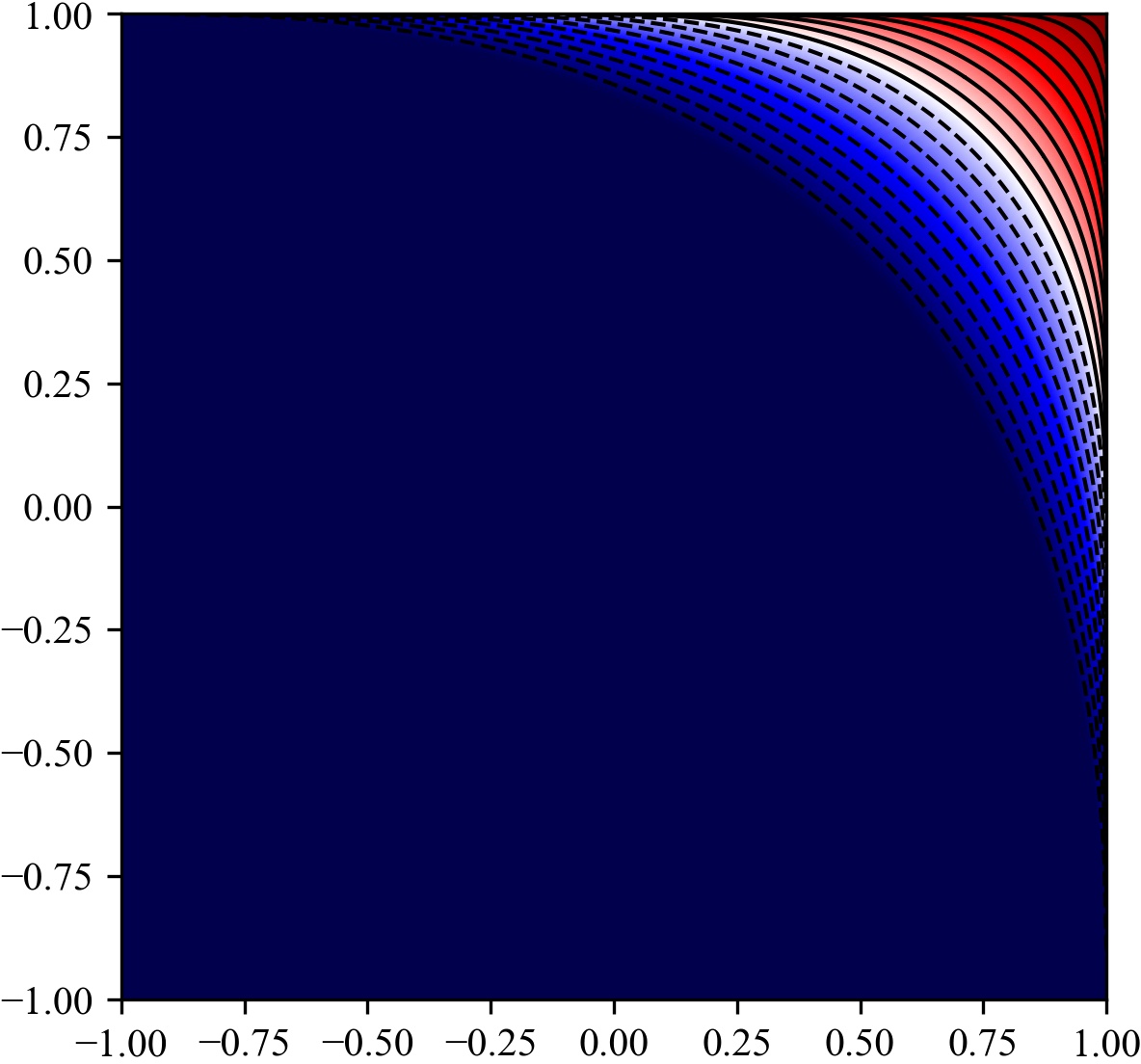}
\caption{Euclidean-based}\label{fig:eucl}
\end{subfigure}
\hfill
\begin{subfigure}{.3\linewidth}\centering
\includegraphics[height=35mm]{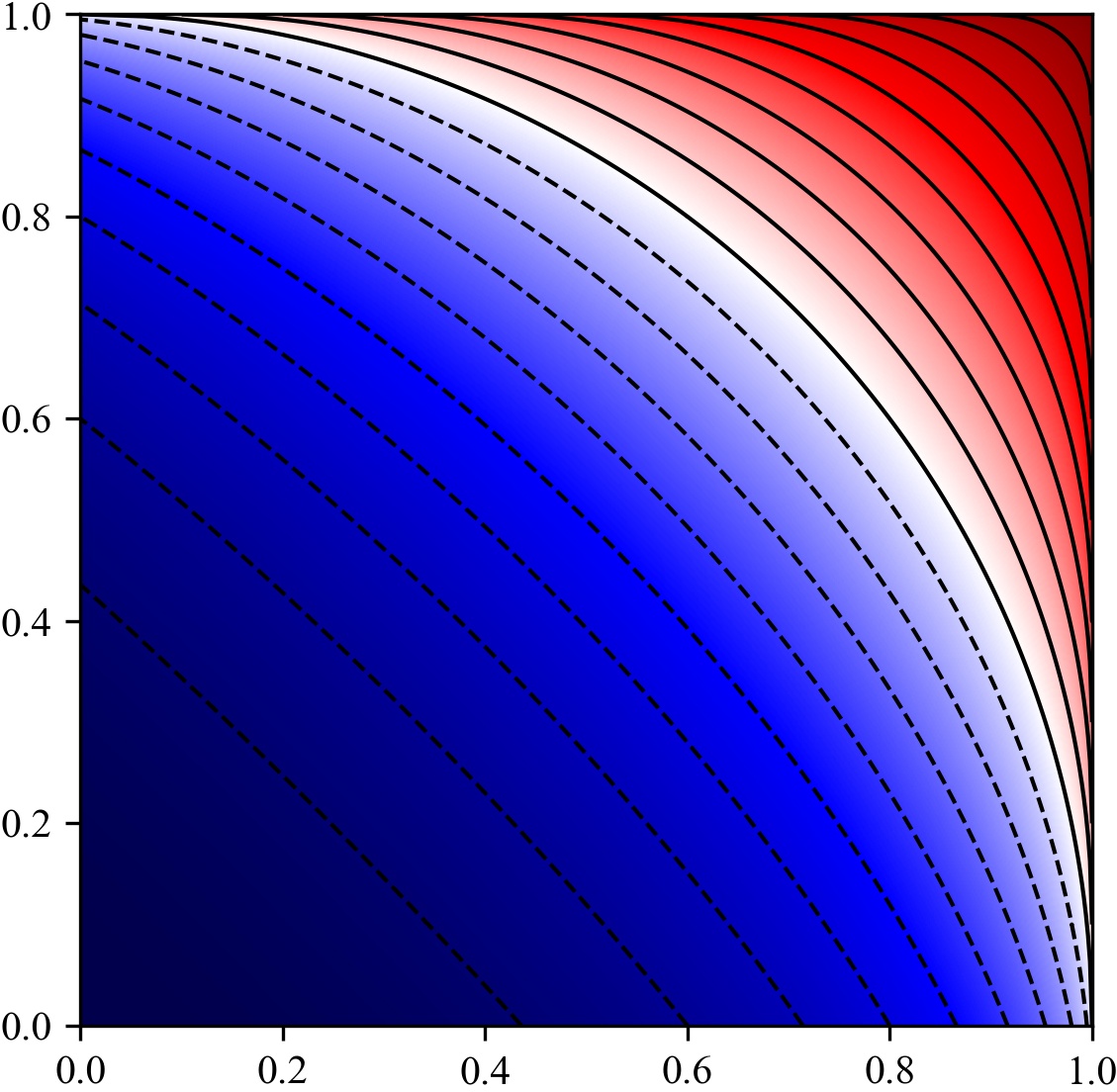}
\caption{Arccos-based}\label{fig:arccos}
\end{subfigure}
\hfill
\begin{subfigure}{.36\linewidth}\centering
\includegraphics[height=35mm]{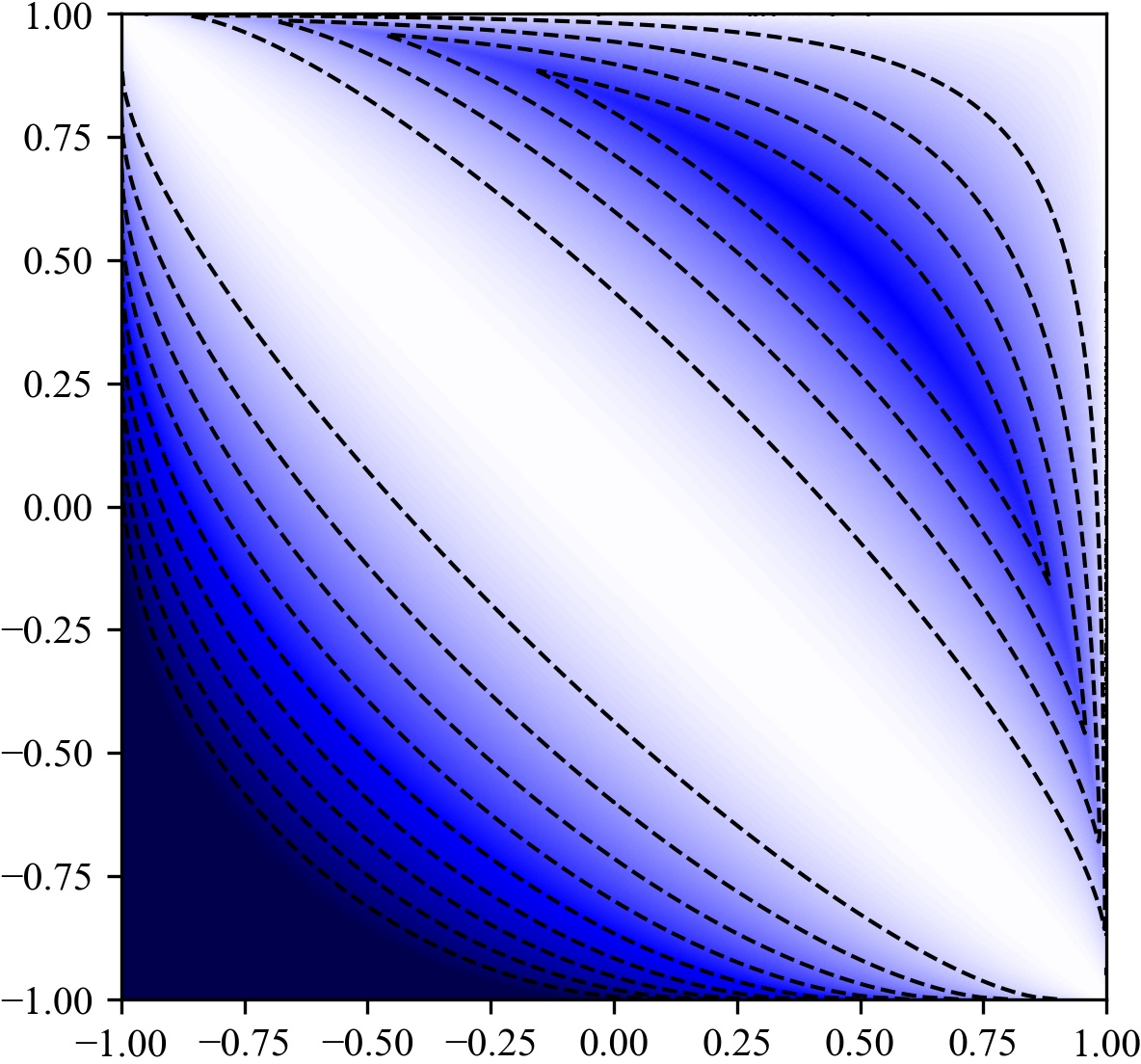}
\caption{Difference}\label{fig:eucl-arccos}
\end{subfigure}
\caption{Euclidean vs.{} Arccos-based triangular inequalities}
\label{fig:triangulars}
\end{figure}

\subsection{Approximation Quality}\label{sec:qual}

In \reffig{fig:triangulars}, we plot the resulting lower bound for the similarity $\cosim(\ve{x},\ve{y})$
given $\cosim(\ve{x},\ve{z})$ and $\cosim(\ve{z},\ve{y})$ using
the Euclidean-based bound (\refeqn{eq:bound-eucl-1}) in \reffig{fig:eucl} and
the Arccos-based bound (\refeqn{eq:bound-acos-1}) in \reffig{fig:arccos}.
A striking difference is visible in the negative domain:
if $\ve{x}$ and $\ve{z}$ are opposite directions, and $\ve{z}$ and $\ve{y}$ are also opposite,
then $\ve{x}$ and $\ve{y}$ must in turn be similar. The Arccos-based bound produces positive bounds here,
while the Euclidean-based bound can go down to $-7$.
But in many cases where we employ Cosine similarity, our data will be restricted to the non-negative domain,
so this is likely not an issue, and could maybe be solved with a simple sign check.
Upon closer inspection, we can observe that the bounds found by the Arccos-based approach
tend to be substantially higher in particular for input similarities around 0.5.
\reffig{fig:eucl-arccos} visualizes the difference between the two bounds.
We can see that the Euclidean bounds are never higher than the Arccos bound, which is unsurprising as the
latter is tight, and the first is a proper lower bound. But we can also see that the difference between the
two (and hence the pruning power) can be as big as 0.5.
This maximum is attained when the input Cosine similarities are 0.5 (i.e., the known angles are $60^\circ$):
The Euclidean bound is -1 then, while the Arccos-based bound is 0.
In the typical use case of Cosine on non-negative values, both bounds are effectively trivial.
However, there still is a substantial difference for larger input similarities.
Averaging over a uniform sampled grid of input values, considering only those where both bounds are non-negative,
the average Euclidean bound is 0.2447, while the average Arccos-based bound is 0.3121, about 27.5\% higher.
Hence, using the Arccos-based bound is likely to yield better performance.

\begin{figure}[tb]\centering
\includegraphics[width=.8\linewidth]{plot/colorbar}
\\
\begin{subfigure}{.32\linewidth}\centering
\includegraphics[width=\linewidth]{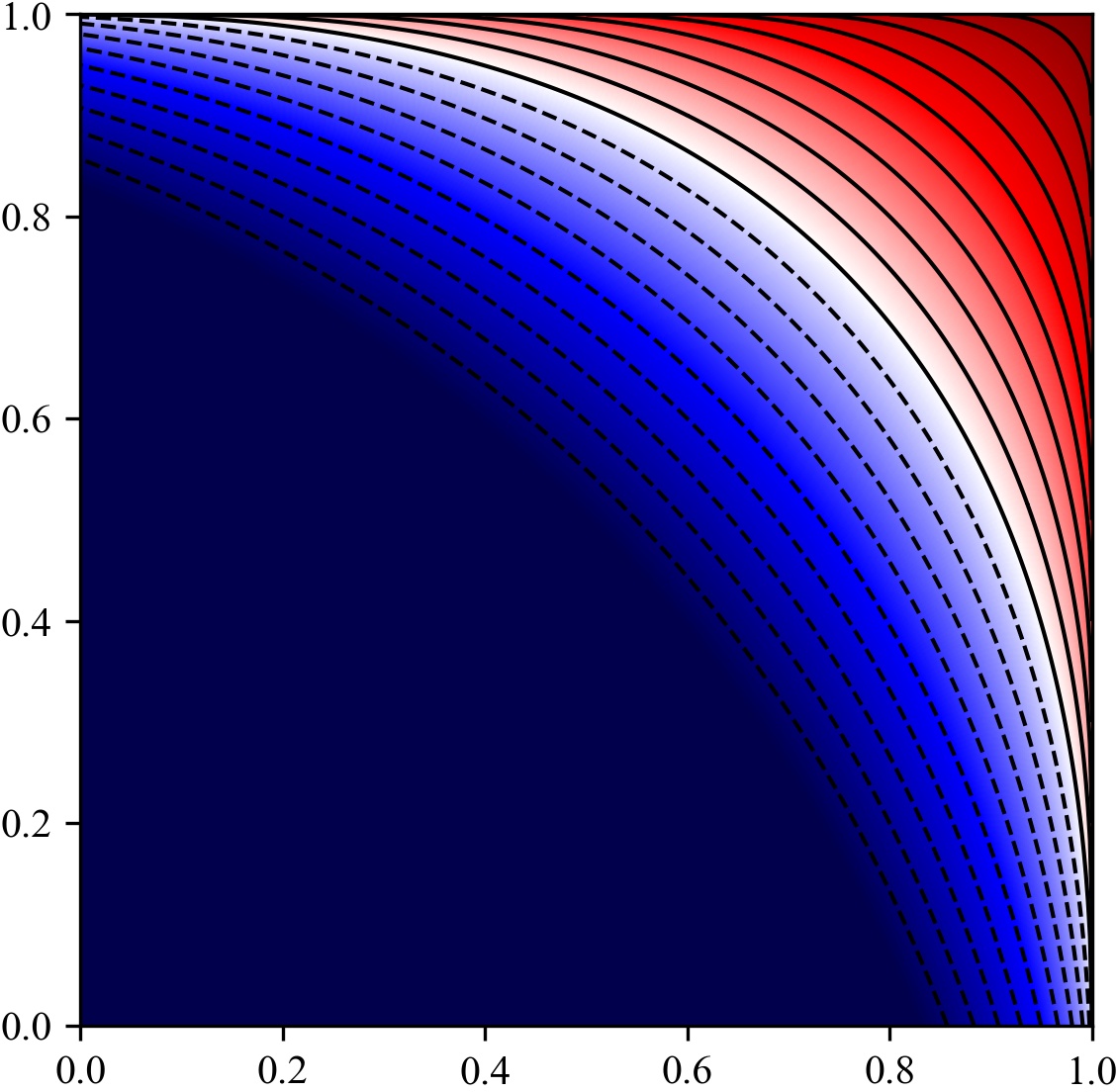}
\caption{Euclidean inequality}\label{fig:b1-eucl}
\end{subfigure}
\hfill
\begin{subfigure}{.32\linewidth}\centering
\includegraphics[width=\linewidth]{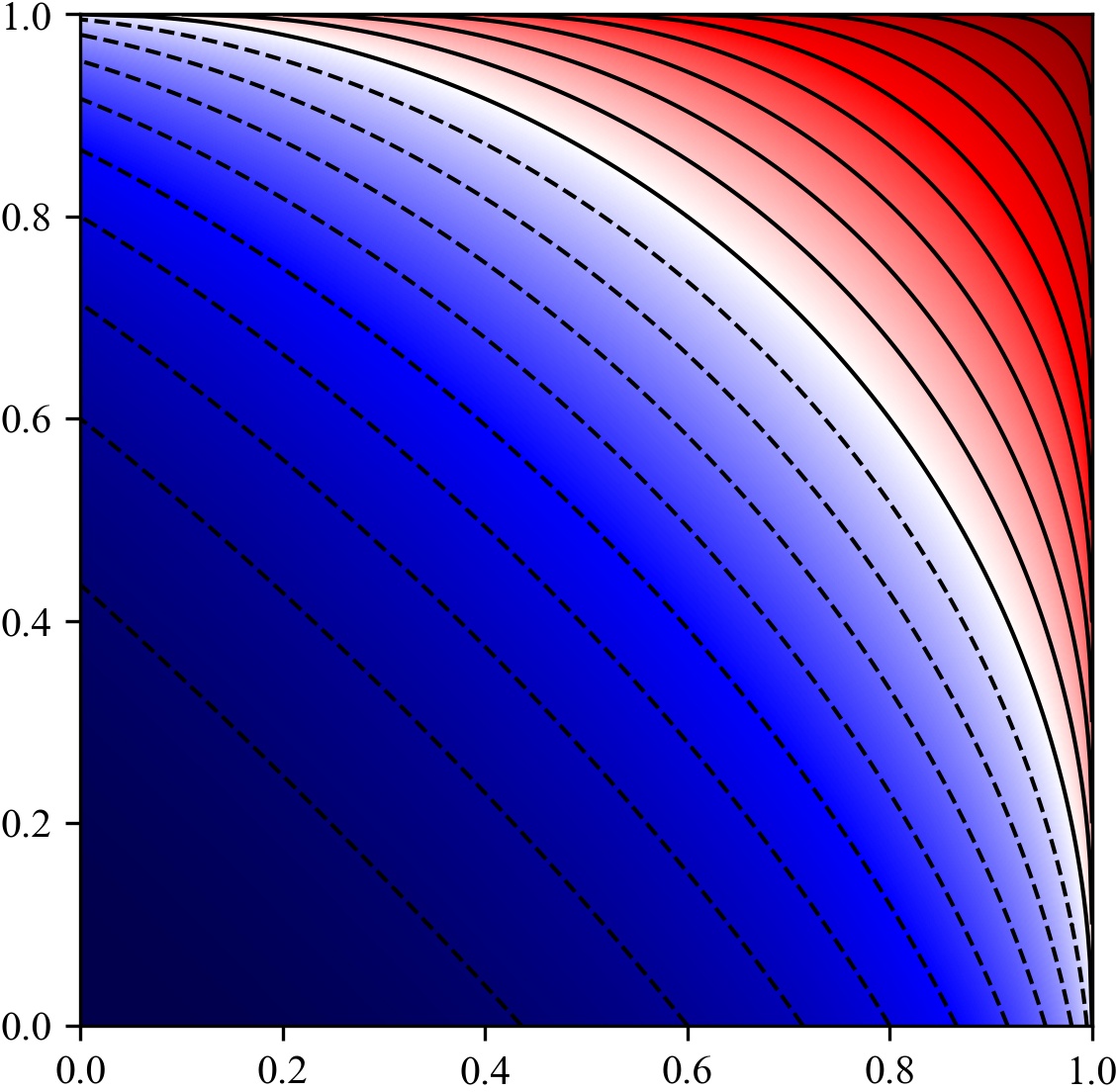}
\caption{Arccos inequality}\label{fig:b1-arccos}
\end{subfigure}
\hfill
\begin{subfigure}{.32\linewidth}\centering
\includegraphics[width=\linewidth]{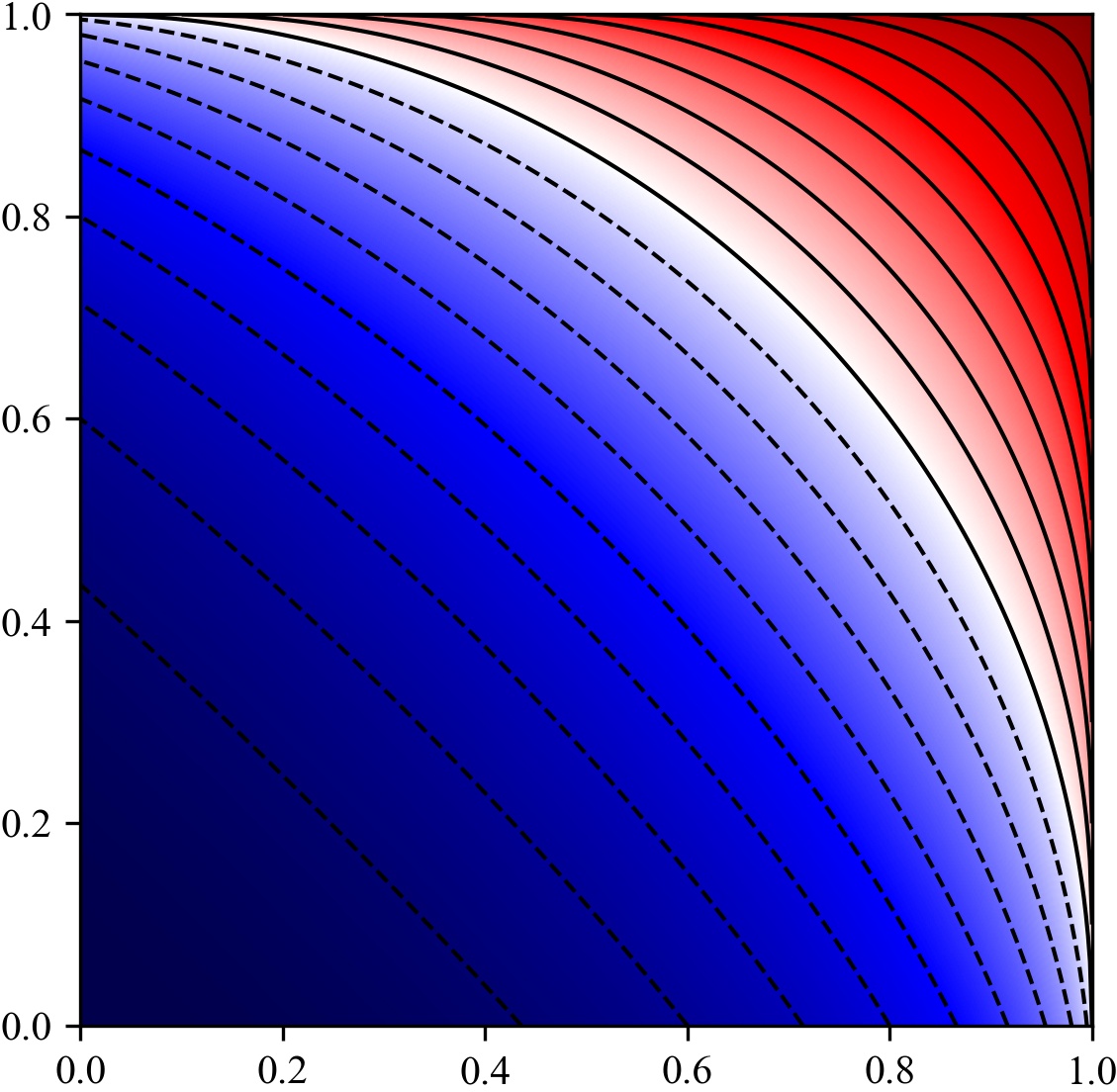}
\caption{Mult inequality}\label{fig:b1-mult}
\end{subfigure}
\\
\begin{subfigure}{.32\linewidth}\centering
\includegraphics[width=\linewidth]{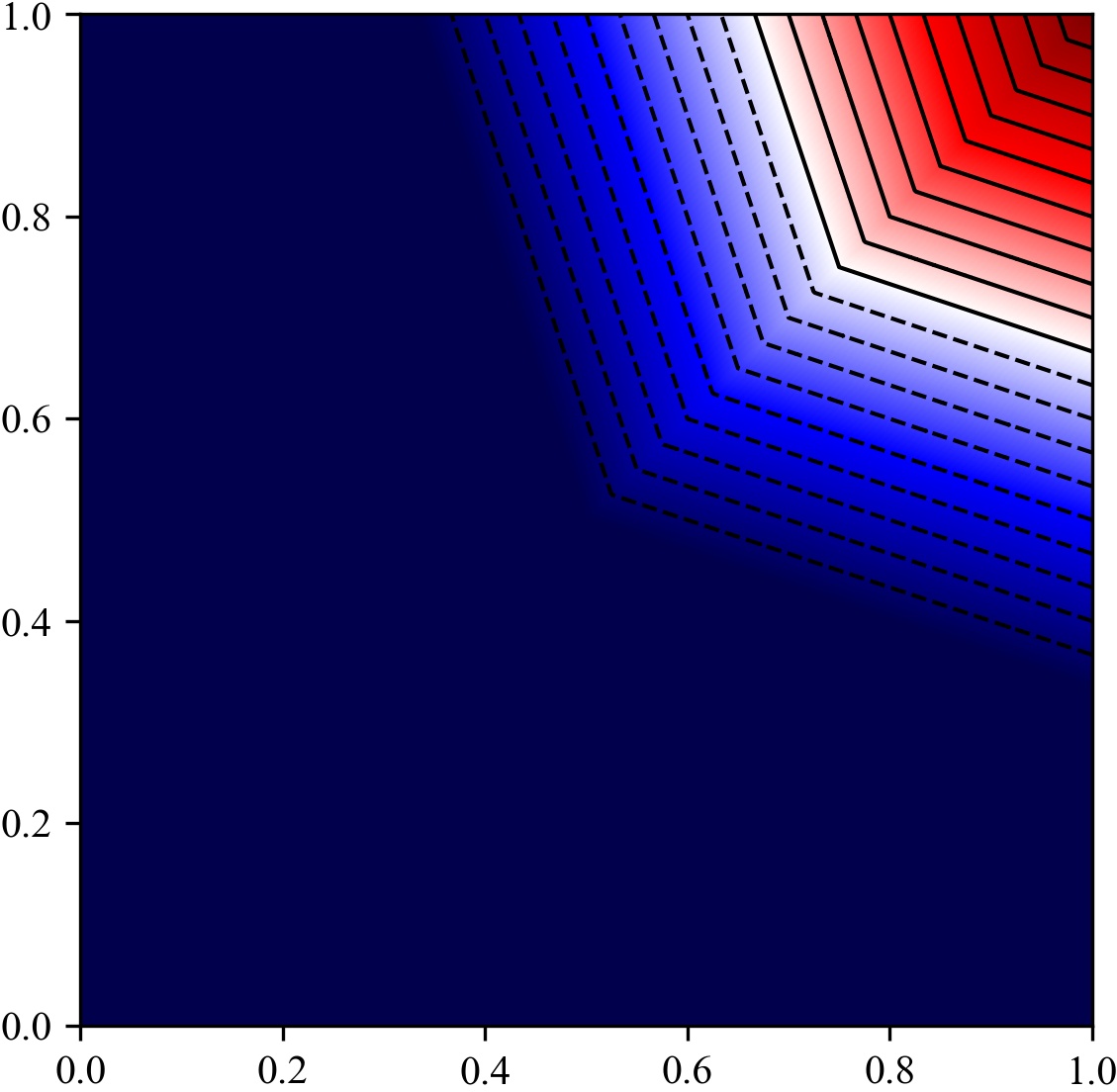}
\caption{Eucl-LB inequality}\label{fig:b1-eucl-lb}
\end{subfigure}
\hfill
\begin{subfigure}{.32\linewidth}\centering
\includegraphics[width=\linewidth]{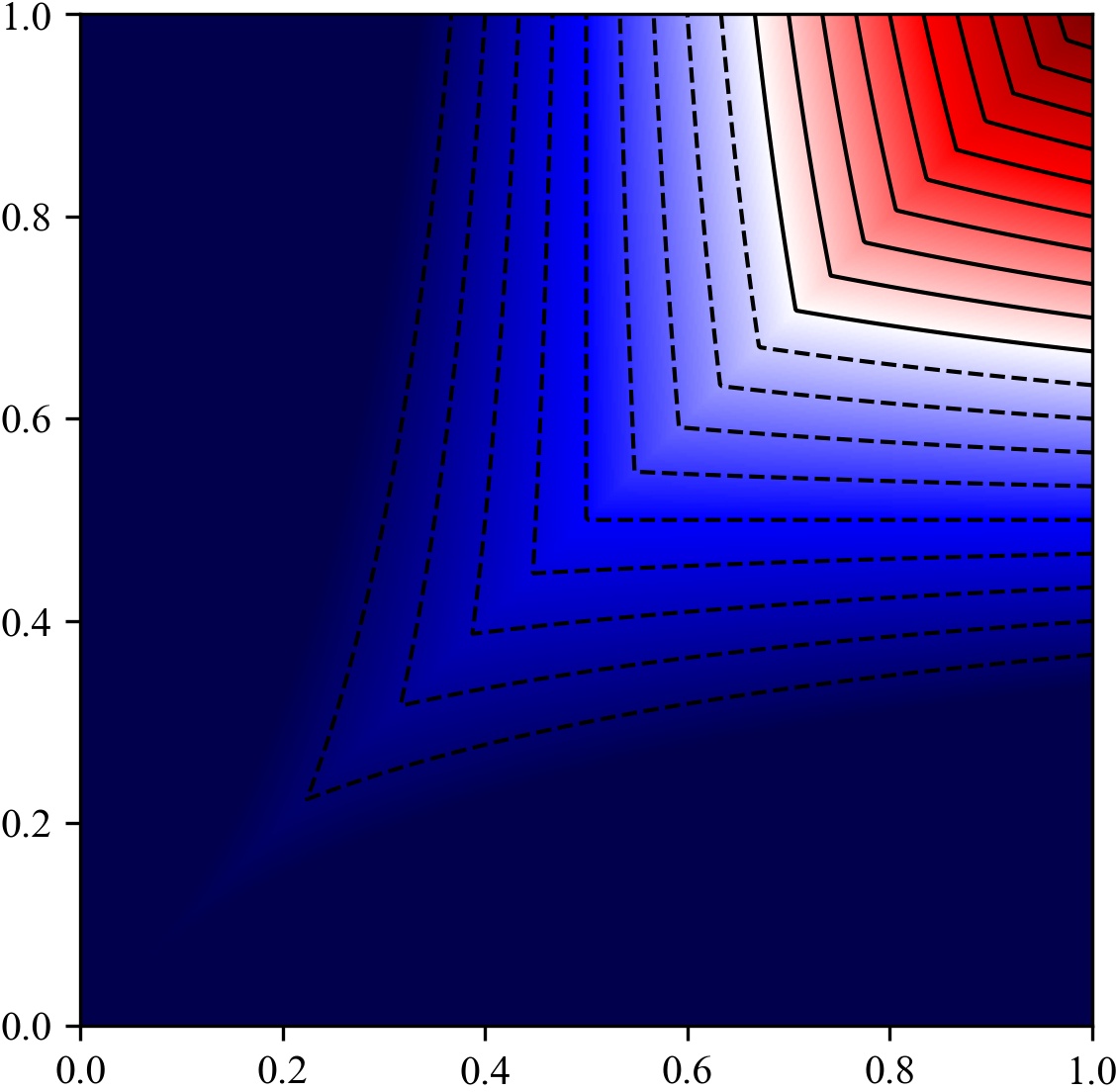}
\caption{Mult-LB2 inequality}\label{fig:b1-mult-lb2}
\end{subfigure}
\hfill
\begin{subfigure}{.32\linewidth}\centering
\includegraphics[width=\linewidth]{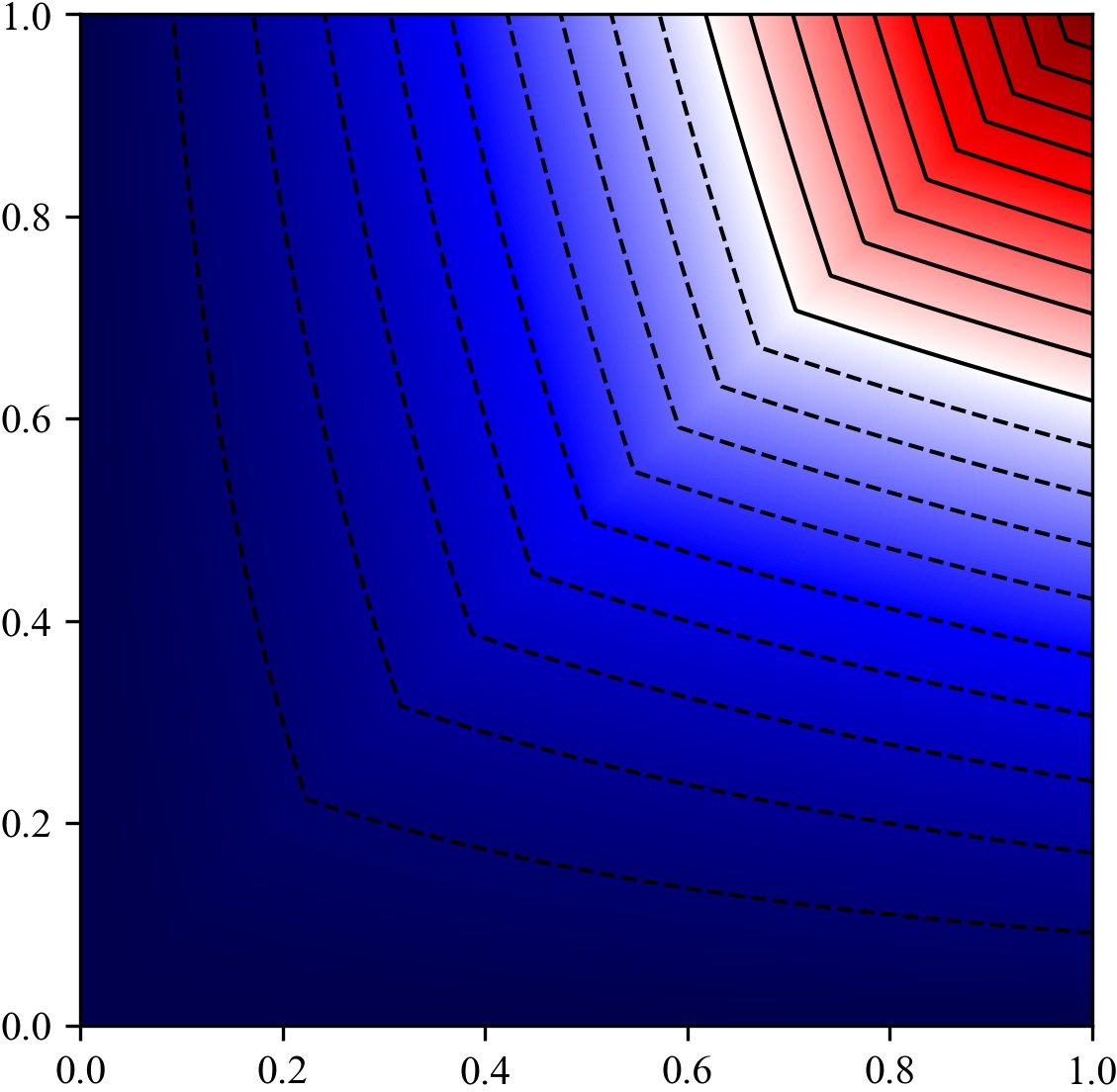}
\caption{Mult-LB1 inequality}\label{fig:b1-mult-lb1}
\end{subfigure}
\caption{Lower bounds for the similarity $\cosim(\ve{x},\ve{y})$ given $\cosim(\ve{x},\ve{z})$ and $\cosim(\ve{z},\ve{y})$
using different inequalities from \reftab{tab:bounds}.}
\label{fig:lower1}
\end{figure}

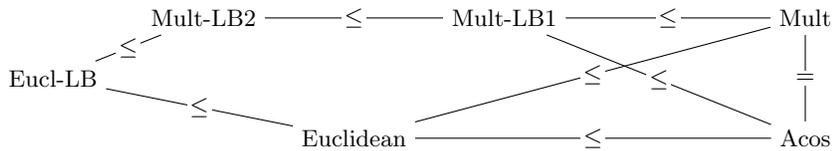
\begin{figure}[tb]
\begin{tikzpicture}[xscale=2,yscale=.8]
\node (eucllb) at (-2,0) {Eucl-LB};
\node (eucl) at (0,-1) {Euclidean};
\node (multlb2) at (-1,1) {Mult-LB2};
\node (multlb1) at (+1,1) {Mult-LB1};
\node (mult) at (+3,1) {Mult};
\node (acos) at (+3,-1) {Acos};
\draw (eucllb) -- (eucl) node[midway, inner sep=1pt, fill=white] {$\leq$};
\draw (eucllb) -- (multlb2) node[midway, inner sep=1pt, fill=white] {$\leq$};
\draw (eucl) -- (acos) node[midway, inner sep=1pt, fill=white] {$\leq$};
\draw (eucl) -- (mult) node[midway, inner sep=1pt, fill=white] {$\leq$};
\draw (acos) -- (mult) node[midway, inner sep=1pt, fill=white] {$=$};
\draw (multlb2) -- (multlb1) node[midway, inner sep=1pt, fill=white] {$\leq$};
\draw (multlb1) -- (acos) node[midway, inner sep=1pt, fill=white] {$\leq$};
\draw (multlb1) -- (mult) node[midway, inner sep=1pt, fill=white] {$\leq$};
\end{tikzpicture}
\caption{Relationships between lower bounds}
\label{fig:relation}
\end{figure}

In the following, we focus on the non-negative domain, to improve the readability of the figures.
In \reffig{fig:lower1}, we show all six bounds from \reftab{tab:bounds}.
\reffig{fig:b1-eucl} is the Euclidean bound,
\reffig{fig:b1-arccos} is the Arccos bound we just saw.
\reffig{fig:b1-mult} is the multiplicative version (\refeqn{eq:bound-acos-2}),
which yields no noticeable difference to the Arccos bound (mathematically, they are equivalent).
\reffig{fig:b1-eucl-lb} is the simplified bound derived from Cosine,
whereas \reffig{fig:b1-mult-lb2} and \reffig{fig:b1-mult-lb1} are the two bounds derived from
the multiplicative version of the Arccos bound.
We observe that Mult-LB1 (\reffig{fig:b1-mult-lb1}) is the best of the lower bounds,
but also that none of the simplified bounds is a very close approximation to the optimal bounds
in the first row.
We obtain the following relationship of the presented lower bounds
(c.f., \reffig{fig:relation}):
\begin{equation*}
\text{Eucl-LB} \leq \text{Euclidean} \leq \text{Arccos} = \text{mult}
\end{equation*}
\begin{equation*}
\text{Eucl-LB} \leq \text{Mult-LB2} \leq \text{Mult-LB1} \leq \text{mult} = \text{Arccos}
\end{equation*}

\begin{figure}[tb]\centering
\begin{subfigure}{.32\linewidth}\centering
\includegraphics[width=\linewidth]{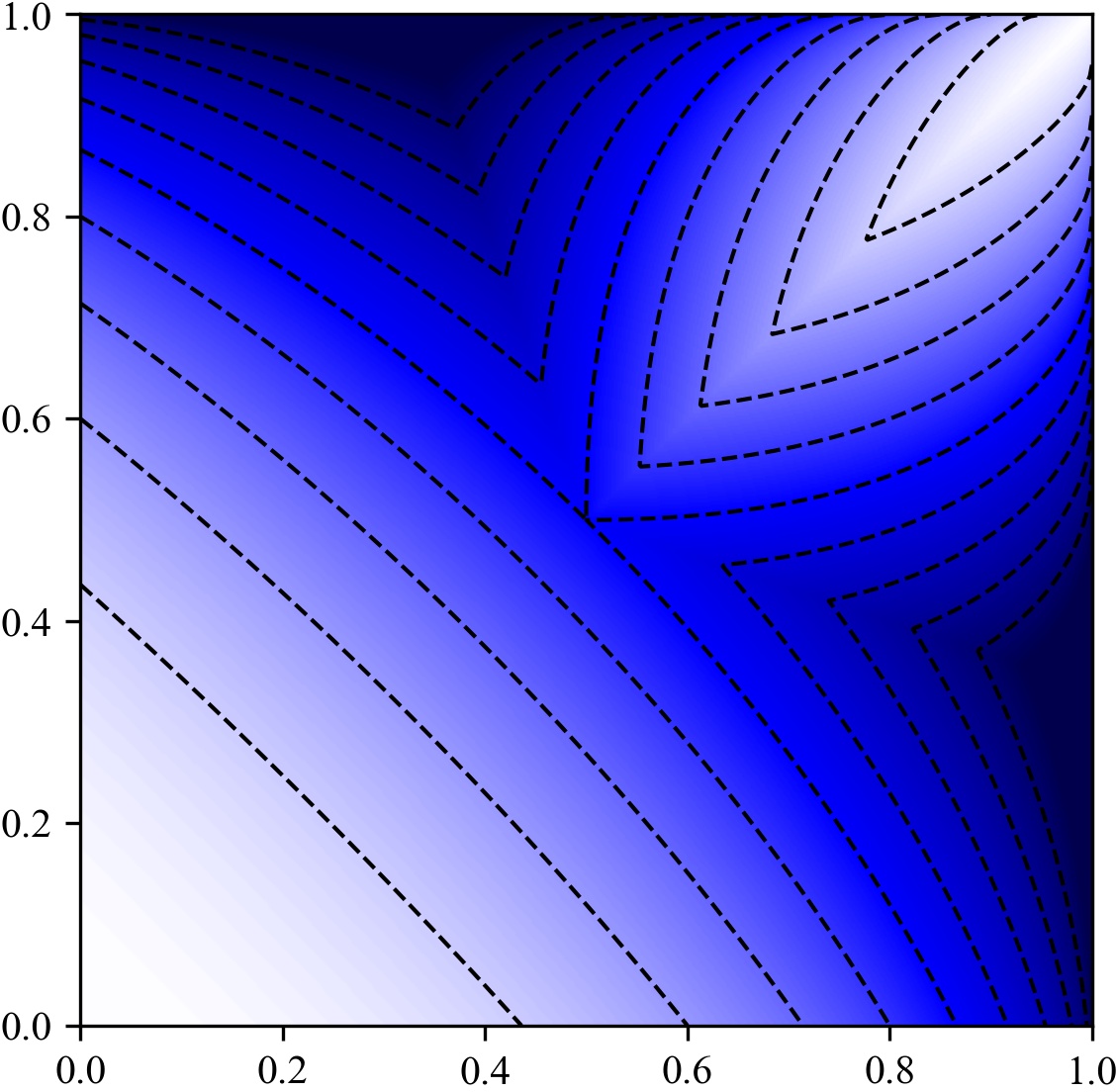}
\caption{Eucl-LB inequality}\label{fig:diff-eucl-lb}
\end{subfigure}
\hfill
\begin{subfigure}{.32\linewidth}\centering
\includegraphics[width=\linewidth]{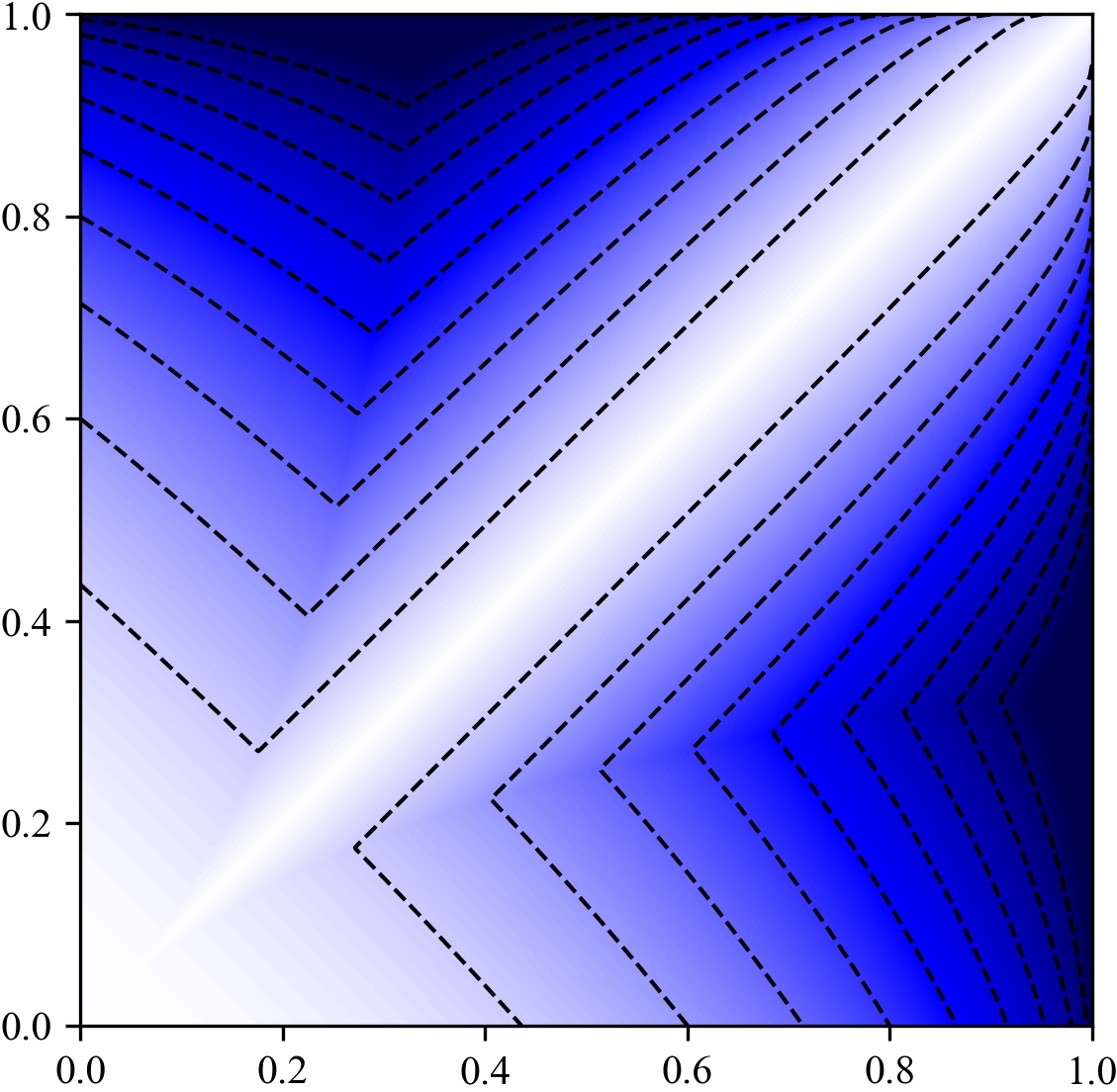}
\caption{Mult-LB2 inequality}\label{fig:diff-mult-lb2}
\end{subfigure}
\hfill
\begin{subfigure}{.32\linewidth}\centering
\includegraphics[width=\linewidth]{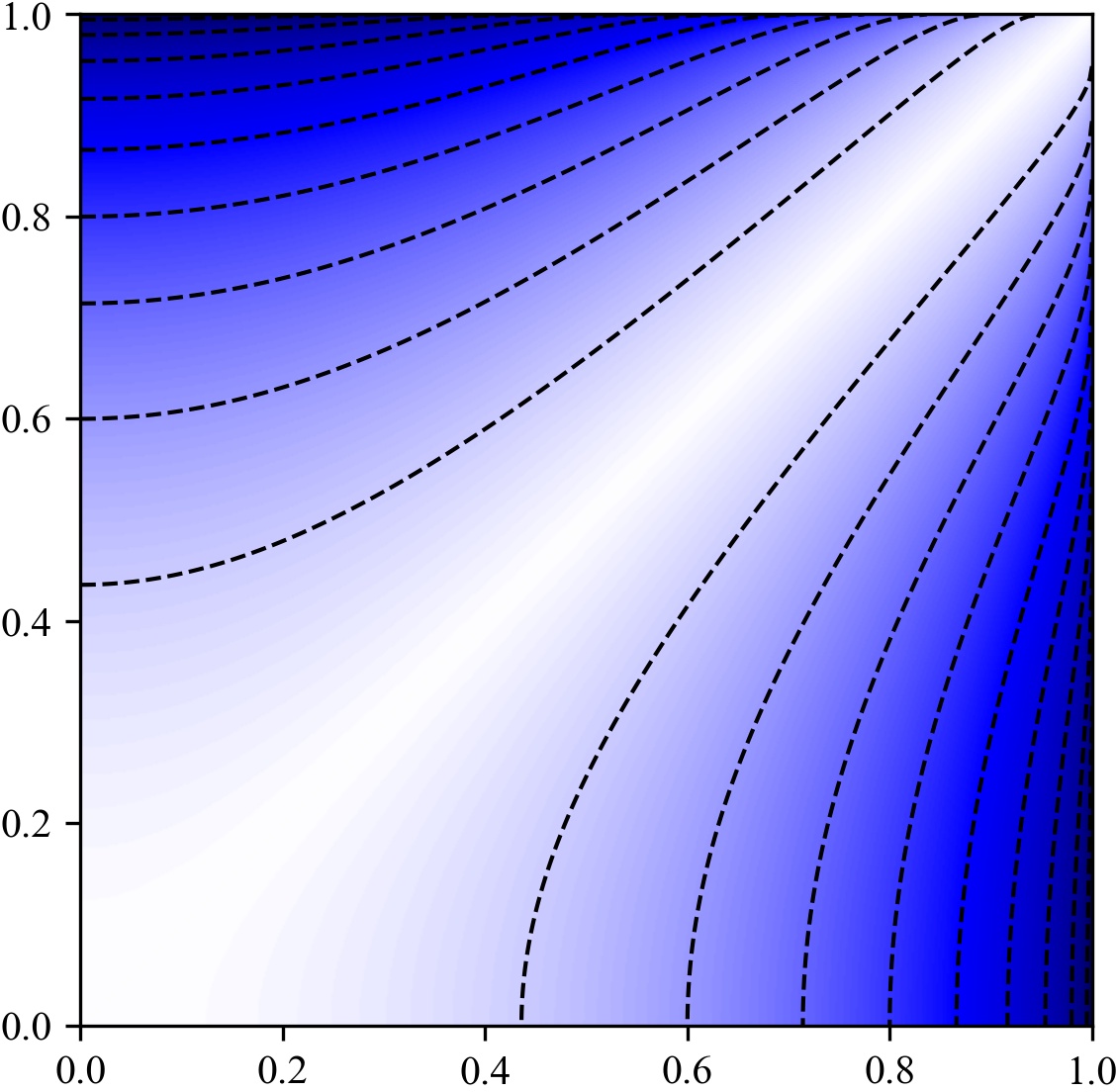}
\caption{Mult-LB1 inequality}\label{fig:diff-mult-lb1}
\end{subfigure}
\\
\includegraphics[width=.8\linewidth]{plot/colorbar}
\caption{Differences between simplified bounds and the tight arccos bound.}
\label{fig:diff}
\end{figure}

In \reffig{fig:diff} we compare the three simplified bounds
(we already compared the Euclidean bound to the tight Arccos bound in \reffig{fig:eucl-arccos}).
While the Mult-LB1 bound is the best of the simplified bounds,
the divergence from the arccos bound can be quite substantial, at least when the two
input similarities are not very close.
As it can be seen from the isolines in the figures (at steps of 0.1),
even if we would consider a bound that is worse by 0.1 or 0.2 acceptable,
there remains a fairly large region of relevant inputs
(e.g., where one similarity is close to 1.0, the other close to 0.8),
where the loss in pruning performance may offset the slightly larger computational
cost of using the Mult bound instead.  

\begin{figure}[tb]\centering
\includegraphics[width=.55\linewidth]{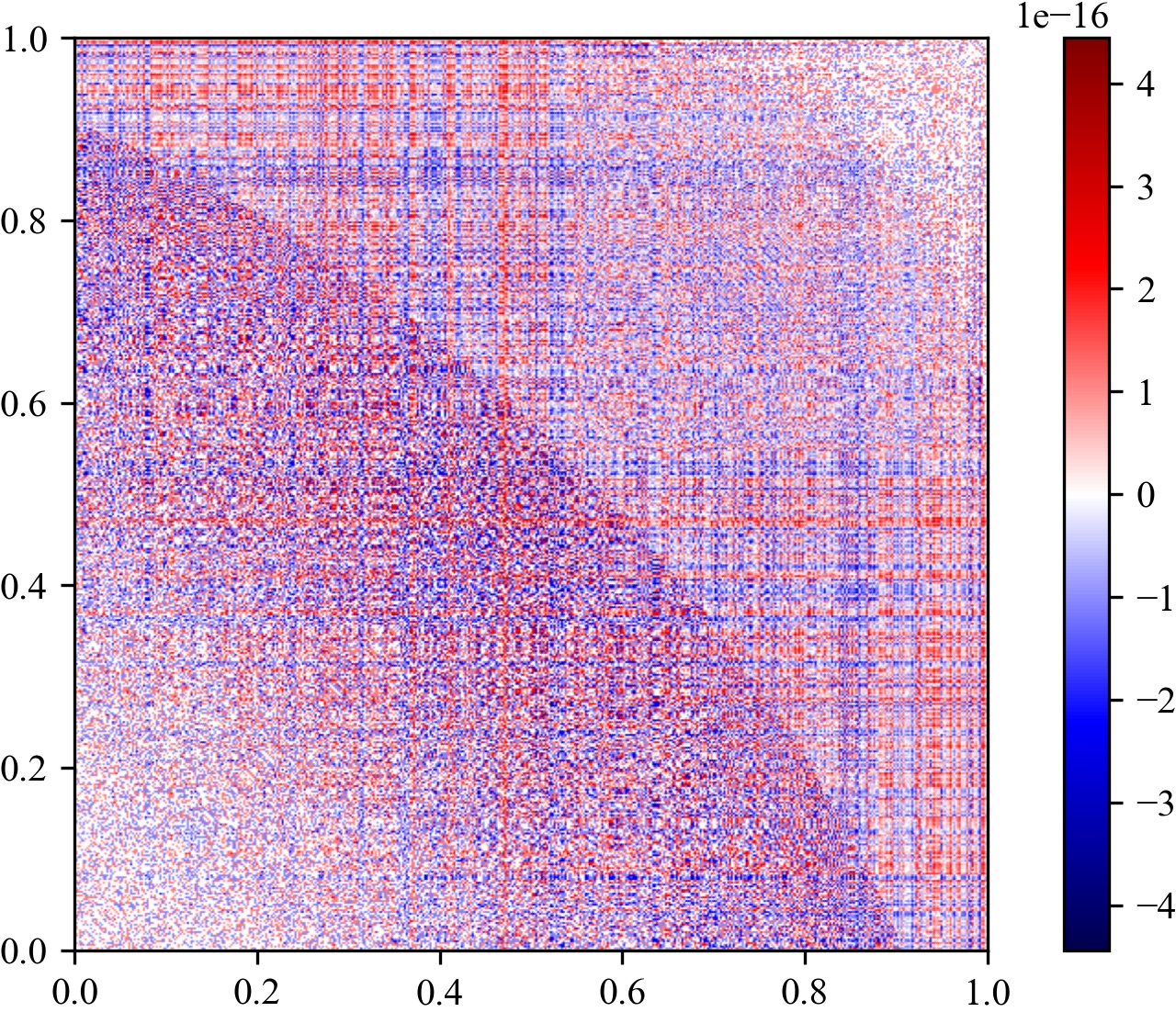}
\caption{Mult inequality}\label{fig:diff-mult}
\end{figure}

\subsection{Numerical Stability}
Mathematically, the Mult bound (\refeqn{eq:bound-acos-2}) is equivalent to
the arccos bound, but more efficient to compute. Given the prior experience
with the numerical problem of catastrophic cancellation, we were concerned that
this equation might be problematic because of the $(1-\cosim^2)$ terms. 
Fortunately, if $\cosim^2\rightarrow 1$, when the problem occurs, the entire
square root will become negligible. We have experimented with some alternatives
(such as expanding the square root to $\sqrt{(1+\cosim(\ve{x},\ve{z}))\cdot (1-\cosim(\ve{x},\ve{z}))\cdot (1+\cosim(\ve{z},\ve{y}))\cdot (1-\cosim(\ve{z},\ve{y}))}$),
but could not find any benefits.
We also compared Mult with the Arccos bound in \reffig{fig:diff-mult}.
While the result appears largely chaotic, the values in this plot are all in the
magnitude of $10^{-16}$, i.e., they are at the expected limit of floating-point
precision. Hence, there does not appear to be a numerical instability in this inequality.

\subsection{Runtime Experiments}
We benchmarked the different equations using Java~11 with double precision floats
and the Java Microbenchmarking Harness JMH~1.32.\footnote{\url{https://openjdk.java.net/projects/code-tools/jmh/}}
The experiments were performed on an Intel i7-8650U using a single thread,
and with the CPU's turbo-boost disabled such that the clock rate is stable at 1.9~GHz
to reduce measurement noise as well as heat effects.
As a baseline, we include a simple add operation to measure
the cost of memory access to a pre-generated array of 2~million random numbers.
Because trigonometric functions are fairly expensive, we also evaluate the JaFaMa library
for fast math as an alternative to the JDK built-ins. JMH is set to perform 5~warmup iterations and
10~measurement iterations of 10~seconds each, to improve the accuracy of our measurements.
We try to follow best practices in Java benchmarking (JMH is a well-suited tool for micro-benchmarking in Java),
but nevertheless, the results with different programming languages (such as C) can be different due to
different compiler optimization, and the usual pitfalls with runtime benchmarks remain~\cite{DBLP:journals/kais/KriegelSZ17}.
\reftab{tab:bench} gives the results of our experiments.
In these experiments, the runtime benefits of the simplified equations are minuscule.
Apparently, the CPU can alleviate the latency of the square root to a large extend (e.g., via pipelining),
and compared to the memory access cost of the baseline operation, the additional 1.6 nanoseconds
will likely not matter for most applications.
The benchmark, however, clearly shows the benefit of the ``Mult'' version over the ``Arccos''
version, which mathematically is equivalent but differs considerably in run time.
While the use of JaFaMa as replacement reduces the runtime considerably, the much simpler ``Mult''
version still wins hands-down and hence is the version we ultimately recommend using.
While ``Mult-LB2'' is marginally faster, it is also much less accurate and hence useful,
as seen in \refsec{sec:qual}.

\begin{table}[bt!]\centering
\caption{Runtime benchmarks for the different equations}
\label{tab:bench}
\setlength{\tabcolsep}{5pt}
\begin{tabular}{lcrlc}
Name & Eq. & Duration & Std.dev. & Accuracy \\
Euclidean & \eqref{eq:bound-eucl-1} &
10.361 ns & $\pm$0.139 ns & \scalebox{.8}{$\bigcirc$}
\\
Eucl-LB & \eqref{eq:bound-eucl-2} &
10.171 ns & $\pm$0.132 ns & $--$
\\
Arccos & \eqref{eq:bound-acos-1} &
610.329 ns & $\pm$3.267 ns & $++$
\\
Arccos (JaFaMa) & \eqref{eq:bound-acos-1} &
58.989 ns & $\pm$0.630 ns & $++$
\\
\bf Mult (recommended) & \eqref{eq:bound-acos-2} &
9.749 ns & $\pm$0.096 ns & $++$
\\
Mult-variant & \footnotemark &
10.485 ns & $\pm$0.022 ns & $++$
\\
Mult-LB1 & \eqref{eq:bound-acos-3} &
10.313 ns & $\pm$0.025 ns & $-$
\\
Mult-LB2 & \eqref{eq:bound-acos-4} &
8.553 ns & $\pm$0.334 ns & $--$
\\
Baseline (sum) & &
8.186 ns & $\pm$0.146 ns & n/a
\end{tabular}
\end{table}
\footnotetext{\refeqn{eq:bound-acos-2} expanded using $(1-x^2)=(1+x)(1-x)$ to obtain the variant\\
$\cosim(\ve{x},\ve{z})\cdot \cosim(\ve{z},\ve{y}) -\sqrt{(1{+}\cosim(\ve{x},\ve{z}))(1{-}\cosim(\ve{x},\ve{z}))(1{+}\cosim(\ve{z},\ve{y}))(1{-}\cosim(\ve{z},\ve{y}))}$}

\vfill
\pagebreak
\section{Conclusions}

In this article, we introduce a triangle inequality for Cosine similarity.
We study different ways of obtaining a triangle inequality, as well as different
attempts at finding an even faster bound.
The experiments show that a mathematically equivalent version of the Arccos-based
bound is the best trade-off of accuracy (as it has optimal accuracy in our experiments)
as well as run-time, where it is only marginally slower than the less accurate alternatives.

Hence, the recommended triangle inequalities for Cosine similarity are:
\begin{align*}
\cosim(\ve{x},\ve{y}) \geq
\cosim(\ve{x},\ve{z})\cdot \cosim(\ve{z},\ve{y}) - \sqrt{(1-\cosim(\ve{x},\ve{z})^2)\cdot (1-\cosim(\ve{z},\ve{y})^2)}
\\
\cosim(\ve{x},\ve{y}) \leq
\cosim(\ve{x},\ve{z})\cdot \cosim(\ve{z},\ve{y}) + \sqrt{(1-\cosim(\ve{x},\ve{z})^2)\cdot (1-\cosim(\ve{z},\ve{y})^2)}
\end{align*}
We can not, however, rule out that there exists a more efficient equation that
could be used instead. As this paper shows, there can be more than one version of the same bound that
performs very differently due to the functions involved.

We hope to spur new research in the domain of accelerating similarity search with metric indexes,
as this equation allows many existing indexes (such as M-trees, VP-trees, cover trees, LAESA, and many more)
to be transformed into an efficient index for Cosine similarity.
Integrating this equation into algorithms will enable the acceleration of data mining algorithms in
various domains, and the use of Cosine similarity directly 
(without having to transform the similarities into distances first) may both allow simplification
as well as optimization of algorithms.
Furthermore, we hope that this research can eventually be transferred to other similarity functions
besides Cosine similarity. We believe it is a valuable insight that the triangle inequality for Cosine
distance contains the product of the existing similarities (but also a non-negligible correction term),
whereas the triangle inequality for distance metrics is additive.
We wonder if there exists a similarity equivalent of the definition of a metric (i.e., a ``simetric''),
with similar axioms but for the dual case of similarity functions, but the results above indicate that we will likely
\emph{not} be able to obtain a much more elegant general formulation of a triangle inequality for similarities.

\bibliographystyle{splncs04}
\bibliography{literature}
\end{document}